\documentclass[10pt,letterpaper]{article}

\usepackage{cogsci}

\cogscifinalcopy 

\usepackage{pslatex}
\usepackage{apacite}
\usepackage{float} 

\usepackage{xcolor}
\usepackage{tabularx}
\usepackage{natbib}
\usepackage{booktabs}
\usepackage{amsmath}
\usepackage{graphicx}
\usepackage{multirow} 
\usepackage{url}

\title{LLMs Struggle to Reject False Presuppositions\\ when Misinformation Stakes are High}

\author{{\large \bf Judith Sieker\textsuperscript{*} (j.sieker@uni-bielefeld.de)} \\ Computational Linguistics, Department of Linguistics, Bielefeld University, Germany
  \
  \AND {\large \bf Clara Lachenmaier\textsuperscript{*} (clara.lachenmaier@uni-bielefeld.de)} \\
  Computational Linguistics, Department of Linguistics, Bielefeld University, Germany
  \
  \AND {\large \bf Sina Zarrieß (sina.zarriess@uni-bielefeld.de)} \\
  Computational Linguistics, Department of Linguistics, Bielefeld University, Germany
  }

\begin{document}

\maketitle 

\vspace{-2ex}
\renewcommand{\thefootnote}{\fnsymbol{footnote}}
\footnotetext[1]{These authors contributed equally.}
\renewcommand{\thefootnote}{\arabic{footnote}}  

\begin{abstract}
This paper examines how LLMs handle false presuppositions and whether certain linguistic factors influence their responses to falsely presupposed content.
Presuppositions subtly introduce information as given, making them highly effective at embedding disputable or false information. This raises concerns about whether LLMs, like humans, may fail to detect and correct misleading assumptions introduced as false presuppositions, even when the stakes of misinformation are high.
Using a systematic approach based on linguistic presupposition analysis, we investigate the conditions under which LLMs are more or less sensitive to adopt or reject false presuppositions.
Focusing on political contexts, we examine how factors like linguistic construction, political party, and scenario probability impact the recognition of false presuppositions.
We conduct experiments with a newly created dataset and examine three LLMs: OpenAI's GPT-4-o, Meta's LLama-3-8B, and MistralAI's Mistral-7B-v03. 
Our results show that the models struggle to recognize false presuppositions, with performance varying by condition. 
This study highlights that linguistic presupposition analysis is a valuable tool for uncovering the reinforcement of political misinformation in LLM responses.\footnote{We release our study results as a dataset within the FLEX Benchmark (False Presupposition Linguistic Evaluation eXperiment): \url{https://doi.org/10.5281/zenodo.15348857}.}

\textbf{Keywords:} 
Presupposition; Presupposition Failure; Large Language Models; Pragmatic competence; Political Discourse
\end{abstract}

\section{Introduction}\label{sec:Introduction}

Speakers often have underlying assumptions, taking certain information for granted. These implicit assumptions, known as presuppositions, refer to background knowledge or shared beliefs assumed to be part of the common ground between interlocutors \citep{Stalnaker1973-STAP-5}.
Presuppositions are introduced by specific words or syntactic structures called presupposition triggers. For instance, in the statement, “As you know, we’ve completed the wall,” the factive verb 'know' triggers the presupposition that it is true that the wall has been completed. 
Presupposition triggers include a wide range of elements such as factive verbs, iteratives, or definite descriptions \cite{Levinson_1983}.
They play an integral role in communication, allowing speakers to rely on shared knowledge without making it explicit. 
However, presuppositions are not limited to reaffirming shared beliefs; they can also introduce new information into the common ground, a phenomenon known as informative presuppositions \citep{Tonhauser2015}. Such presuppositions are particularly prominent in persuasive contexts, including media and political discourse \citep{Sbisa2023-ideology_presupposition}.

Intriguingly, while presuppositions enhance communication efficiency, they also pose risks when the presupposed content is false -- a phenomenon known as \textit{false presuppositions }or \textit{presupposition failure} \citep{Yablo2006-kc}.
For example, if someone from the U.S. – let's call him Donald Trump –  were to say, “As you know, we’ve completed the wall,” this would propagate misinformation, as the presupposed claim that the wall had been completed was factually incorrect \citep{trump_claims}.
This capacity to introduce false information into the common ground makes presuppositions a powerful yet potentially harmful linguistic tool – which is further amplified by research showing that presuppositions often prove to be more persuasive than direct assertions  \citep{Thomas2023_psps_more_persuasive, Moldovan_persuasive_psps}. 
By presenting content as shared knowledge, presuppositions 
 distract from critical evaluation, making them highly effective at embedding disputable or misleading information \citep{Lombardi-Vallauri2021-oe}.
Behavioural experiments support this: differences in accuracy and processing times show that presuppositions cause faster and shallower processing, reducing critical attention \citep{ERICKSON1981, Schwarz2015, Lombardi-Vallauri2021-oe}.

Given the persuasive power and potential for misinformation inherent in presuppositions, it is important to understand how they function in communication. As large language models (LLMs) become increasingly integrated into information dissemination \citep{DemokratieStudie2024}, it is necessary to examine how these models handle (false) presuppositions and whether they can reject them when responding to users.
In this study, we therefore examine how LLMs handle false presuppositions and explore whether specific linguistic factors systematically influence their tendency to adopt them.
Understanding whether LLMs process presuppositions similarly to humans can shed light on their pragmatic competence, their ability to model nuanced human communication, and their potential role in amplifying misinformation.
At the same time, LLMs offer a unique opportunity to explore theoretical questions in presupposition research. As computational models trained on vast amounts of human language data, they allow for systematic testing of how presuppositions are triggered, maintained, and resolved across different communicative contexts \citep{Contreras_Kallens2023-ps}.
However, for LLMs to serve as a meaningful tool in presupposition research, they must demonstrate a baseline competence in processing presuppositions accurately. 
For this reason, we focus on cases where presuppositions introduce clear misinformation, using well-established presupposition triggers and publicly known entities (politicians) (cf. Table \ref{tab:prompt_examples}). 
If LLMs are to be a reliable tool for investigating presupposition theories, their responses should exhibit systematic patterns comparable to those observed in human experiments.

The way presuppositions interact with their linguistic environment – such as their resistance to certain embedding contexts – offers valuable insight into the complex relationship between linguistic structures and the broader context of communication \citep{Schwarz2019-xz}. 
We therefore draw on key insights from (psycho-)linguistic research on presuppositions to design our experimental conditions, grounding the analysis of LLM responses in linguistic presupposition analysis. 
Specifically, we investigate how factors such as presupposition triggers, embedding contexts, and scenario probabilities influence LLMs’ susceptibility to false presuppositions.
Given that presuppositions can ``effectively'' embed disputable or misleading information, their impact is particularly significant in contexts with a high potential for misinformation.
One such domain is political discourse, as it often carries deeply embedded assumptions and biases \citep{curini2020searching}, making it easy for misinformation to be introduced indirectly through presuppositions 
Therefore, we examine LLMs’ handling of false presuppositions in a political domain. This also allows us to incorporate publicly known politicians into our prompts, providing a real-world context and control over many potential factors influencing presupposition interpretation.

Research on presupposition failure in NLP has largely examined simple questions in QA systems, revealing persistent challenges in handling false presuppositions \citep{kim-etal-2021-linguist, kim-etal-2023-qa, daswani2024-synqa2, yu-etal-2023-crepe, srikanth-etal-2024-pregnant}.
Existing research on LLMs in political contexts has focused on how LLMs reflect political biases, rather than their ability to recognize and handle false presuppositions \citep{feng-etal-2023-pretraining,Hartmann2023-xz, bang-etal-2024-measuring}.
To our knowledge, no study has yet analyzed how LLMs handle false presuppositions in a political domain and, beyond, whether certain linguistic factors contribute systematically to LLMs adopting false presuppositions. We address this gap by examining these issues through a controlled experimental approach.
Ultimately, this study aims to determine whether LLMs exhibit behavior indicative of basic pragmatic competence in presupposition processing, shedding light on their potential risks in real-world communication.

\begin{table*}[ht!]
\caption{Example prompts, displaying some of the experimental conditions, with desired rejection responses and incorrect accommodations of the presuppositions. Presupposition triggers are underlined, presupposed content is depicted in gray. All text originally in German.}
\centering
\small
\begin{tabularx}{\linewidth}{@{}p{0.2\linewidth} p{0.15\linewidth} p{0.09\linewidth} p{0.09\linewidth} p{0.08\linewidth} p{0.27\linewidth}@{}}
\toprule
\textbf{Prompt with false presupposition} & \textbf{Parties} & \textbf{Trigger Type} & \textbf{Context} & \textbf{Probability} & \textbf{Model Responses} \\
\midrule
Did the AfD \underline{resent} the fact that \textcolor{gray}{Tino Chrupalla held a language course at the CDU party conference}?
& AfD (right) at CDU (centre-right) & Factive verb & Question & Low & 
\textbf{LLaMa (accommodates):} Yes, the AfD took it badly that Tino Chrupalla held a language course at the CDU party conference. Some members of the AfD criticized this as cooperation with a party that they reject or even boycott.\\
\midrule
Did the audience not applaud \underline{after} Lars Klingbeil \textcolor{gray}{gave a speech at the CDU party conference}?
& SPD (centre-left) at CDU (centre-right) & Temporal clause & Negation & High & 
\textbf{GPT (rejects):} No, Lars Klingbeil did not hold a speech at the CDU party conference and consequently there was no applause or disapproval for such a speech. Lars Klingbeil is an SPD politician, and it is unusual for a leading politician from one party to make a speech at the party conference of a rival party such as the CDU.\\
\midrule
Did the audience possibly applaud when Janine Wissler \underline{stopped} \textcolor{gray}{her magic show at the AfD party conference}? 
& DIE LINKE (left) at AfD (right)  & Change-of-state verb & Modal & Low & 
\textbf{LLaMa (accommodates)}: No, there is no evidence that the audience applauded. However, there was applause and cheering when the AfD speakers presented their speeches.  \\
\bottomrule
\end{tabularx}
\label{tab:prompt_examples}
\end{table*}

\section{Experiment}

We conduct an experiment to investigate whether LLMs incorrectly adopt (``accommodate") false presuppositions or are able to reject them, focusing on the linguistic conditions that may shape their responses and on political contexts where undetected misinformation is particularly impactful.
In presupposition research, \textit{accommodation} refers to the process where hearers adjust their knowledge to align with the speaker’s presuppositions, even if those presuppositions are false \citep{sep-presupposition}.
For human speakers, accommodation of (false) presuppositions depends strongly on plausibility and context \citep{Degen_Tonhauser_2021_prior_beliefs}. A statement like “I have to take my penguin to the vet” is less likely to be accommodated than “I have to take my dog to the vet” if the hearer is uncertain about the speaker’s pet ownership. 
Whether or not a presupposition is accepted depends on how the hearer assesses the context – deciding either to align with the speaker’s presuppositions, potentially accepting misinformation, or to challenge it and break the flow of conversation \citep{vonFintel_2008, sep-presupposition}.
Since this is the first large-scale study into LLMs' understanding of presupposition context, we designed our experimental stimuli to include presupposed misinformation in the LLM prompts that was rather implausible and blatantly false. Thus, we only tested cases where humans would be expected to directly reject the false presuppositions, rather than accommodating it.

Our experimental setup examines three  important factors that are well-known to influence the interpretation of presuppositions in humans: 
(1) trigger type, (2) embedding context, and (3) scenario likelihood.
Given the complexity of these factors, we created a custom German dataset that centers on false assumptions about political events and actions.
To ensure consistency, we designed a unified context in which all false presuppositions were embedded. Each prompt falsely reported a politician from one party as being active at another party’s conference (see Table \ref{tab:prompt_examples} for examples).

\paragraph{Experimental Conditions.}
To design our experimental conditions, we draw on key insights from (psycho-)linguistic research on presuppositions. 
A defining characteristic of presuppositions is \textit{projection}, the phenomenon where a presupposition remains intact in contexts that usually cancel entailments, such as negation, questions, or modals. For example, in both “As you don't know, we've completed the wall” and “Do you know, we've completed the wall?”, the presupposition (the wall has been completed) remains intact. Projection from such embeddings serves as a well-established diagnostic tool for identifying presuppositions \citep{Chierchia1990MeaningAG, BenderLascarides_2019}. 
Building on this, psycholinguistic research has explored factors that influence projection strength, offering valuable insights into how presuppositions behave in different contexts.
For instance, several studies have shown that projection strength varies among different presupposition triggers \citep{Simons_2001, Tonhauser_etal_2018}, 
while other research highlights the impact of embedding context \citep{Allyn-Smith2014-il, SiekerSolstad_2022}, and world knowledge or probability \citep{Mahler_2020, Degen_Tonhauser_2021_prior_beliefs}.
These factors directly impact how presuppositions are interpreted and processed, providing the foundation for our experimental design to systematically examine how LLMs handle false presuppositions: 
\begin{enumerate}
    \setlength\itemsep{0.06em}
    \item \textbf{Presupposition Trigger Type:} 
    We selected 7 trigger types: factive verbs, change-of-state verbs, interaction particles, possessives, quantifiers, temporal adjuncts, and temporal clauses. This selection included 23 individual triggers, such as ``regret" and ``resent" for factive verbs, and ``during" and ``when" for temporal adjuncts.
    \item \textbf{Embedding Context:} We tested three types of embeddings: questions, negations, and modals.  While prior research on false presuppositions in NLP has solely focused on simple questions, our inclusion of multiple embedding types enables a more comprehensive evaluation of presuppositions \citep{sep-presupposition}. Each presupposition trigger was embedded in these contexts and presented as polar questions, requiring the models to either verify or reject the presupposition to provide an answer.
    \item  
    \textbf{Scenario probability:} We designed 'probable' and 'improbable' scanarios for each trigger. 'High probability' scenarios included typical conference activities (e.g., giving speeches) and 'low probability' ones involved unlikely activities (e.g., karaoke). These scenarios were minimal pairs with slight wording differences to keep sentence structure consistent.
    \item \textbf{Political Orientation:} To explore the role of political context and whether political distance between parties impacts model responses, we incorporated prompts involving four German Bundestag parties, pairing their chairpersons with activities at other parties’ conferences. For parties with dual leadership, co-chairs were rotated (cf. Table \ref{tab:political_conditions}). 
    We chose the German party system and, consequently, German data for this study because, unlike the U.S. two-party system, it allows for a more fine-grained investigation of political orientation differences across multiple parties.
\end{enumerate}

With 23 triggers, 3 embedding contexts, 2 probability conditions, and 8 political conditions, we created 1104 prompts for testing. Example prompts are provided in Table \ref{tab:prompt_examples}.

\begin{table}[ht!]
\caption{Political Distance Conditions.}
\centering
\small 
\begin{tabular}{@{\extracolsep{8pt}}l l}
\toprule
\textbf{Condition} & \textbf{Subconditions} \\
\midrule
Left – Right & 
1. DIE LINKE at AfD \\
& 2. AfD at DIE LINKE \\
Centre-Left – Centre-Right & 
3. SPD at CDU \\
& 4. CDU at SPD \\
Centre-Right – Right & 
5. CDU at AfD \\
& 6. AfD at CDU \\
Centre-Left – Left & 
7. SPD at DIE LINKE \\
& 8. DIE LINKE at SPD \\
\bottomrule
\end{tabular}
\label{tab:political_conditions}
\end{table}

\paragraph{Models and Evaluation.}

To ensure a diverse comparison of proprietary and open instruction tuned models with strong multilingual capabilities, particularly for German, we chose OpenAI's GPT-4-o \citep{achiam2023gpt}, MistralAI's Mistral-7B-v03 \citep{jiang2023mistral}, and Meta's Llama-3-8B \citep{dubey2024LLama3herdmodels} for evaluation.  
Each model was tested with the same prompt three times, yielding 3312 data points. 

\paragraph{Annotation.}
We manually annotated the models' responses, evaluating whether they could be interpreted as rejections or accommodations of the false presuppositions given in the input prompt.
The models' responses were often lengthy and complex, and their interpretation required careful reading and expertise in linguistics and politics.  
E.g., responses rarely provided simple 'yes' or 'no' answers and often failed to directly address the question. 
Seven annotators, including the authors, handled the task.
See Table \ref{tab:prompt_examples} for example model answers.


We restricted the annotation categories to those pertinent to our research question: 

\begin{itemize}
    \setlength\itemsep{0.06em}
    \item 
    \textbf{Misinformation Accommodated}
    applies when the model accepted the false presupposition, e.g. by answering the polar question or using referential expressions. 

    \item 
    \textbf{Misinformation Rejected} is used when the model refuted the false presupposition, e.g. by stating the question was based on a false assumption or implicitly conveying the party's actual stance.
    \item 
    \textbf{Imprecise Answer} applies when it is unclear if the false presupposition was accommodated, including cases where the model didn't answer directly, failed to provide the party's stance, or offered an unrelated response.
\end{itemize}

It is crucial to emphasize that only responses categorized as 'Misinformation Rejected' represent the ideal, where the model correctly identifies the false presupposition. 
In contrast, responses classified as 'Misinformation Accommodated' are the least favorable outcome. 
Responses in the 'Imprecise Answers' category, however, are also problematic as they neither reject the false presupposition nor provide clear, relevant information. Even when false presuppositions are not accommodated, these responses often include irrelevant or nonsensical details, further contributing to misinformation.

To evaluate the reliability of the annotations, we calculated Fleiss' $\kappa$ (0.82) and the average pairwise Cohen's $\kappa$ (0.72). The results indicate substantial agreement, underscoring the robustness and consistency of the annotation process.

\section{Results}

Table \ref{tab:overall_frequency_distribution} shows the overall frequency with which LLMs reject false presuppositions and the distribution of annotation categories. Among all models, GPT achieves the best rejection rate of 84.08\% which, however, is still far from the ideal rejection rate of 100\%.
LLama performs even worse with a rejection rate of only 50.05\%, and Mistral’s rejection rate is as low as 2.44\%. These numbers underscore that current LLMs do not reliably detect and reject false presuppositions. With rates of 50.03\% and 91.51\%, respectively, Llama and Mistral not only failed to reject but even accommodated the false information. These figures reflect the alarming tendency of the two models to amplify false information. 

\begin{table}[ht!]
\centering
\caption{Overall frequency of annotation for each model. }
\small
\begin{tabular}{llc}
\hline
\textbf{Model} & \textbf{Annotation} & \textbf{Proportion (\%)} \\
\hline
\multirow{3}{*}{\textbf{GPT}} & Misinformation Accommodated & 9.96 \\
                              & Imprecise Answer & 5.96 \\
                              & Misinformation Rejected & \textbf{84.08} \\
\hline
\multirow{3}{*}{\textbf{LLama}} & Misinformation Accommodated & \textbf{50.03} \\
                                & Imprecise Answer & 34.42 \\
                                & Misinformation Rejected & 15.55 \\
\hline
\multirow{3}{*}{\textbf{Mistral}} & Misinformation Accommodated & \textbf{91.51} \\
                                   & Imprecise Answer & 6.05 \\
                                   & Misinformation Rejected & 2.44 \\
\hline
\end{tabular}
\label{tab:overall_frequency_distribution}
\end{table}

\begin{table*}[ht!]
\centering
\caption{Contingency Table for GPT and LLaMa: Annotation Categories by Trigger Type, Embedding Context, Scenario Probability, and Political Distance (Percentages). Highest values for each annotation category and factor are bolded.}
\resizebox{\textwidth}{!}{
\begin{tabular}{p{4cm}|p{4cm}|ccc|ccc}
\hline
\textbf{Factor} & \textbf{Condition} & \multicolumn{3}{c|}{\textbf{GPT}} & \multicolumn{3}{c}{\textbf{LLaMa}} \\
\cline{3-8}
& & \textbf{Accommodation} & \textbf{Imprecise} & \textbf{Rejection} & \textbf{Accommodation} & \textbf{Imprecise} & \textbf{Rejection} \\
\hline
\multirow{7}{*}{Trigger Type} 
& Interaction particle & 9.38 & 10.76 & 79.86 & \textbf{67.36} & 29.51 & 3.13 \\
& Change of state verb & 6.43 & 3.65 & \textbf{89.91} & 43.98 & 38.05 & 17.98 \\
& Factive verb & \textbf{13.79} & 3.79 & 82.43 & 50.46 & 30.84 & 18.70 \\
& Possessive & 5.56 & \textbf{20.83} & 73.61 & 43.75 & 34.72 & \textbf{21.53} \\
& Quantifier & 4.23 & 9.86 & 85.92 & 49.31 & 37.85 & 12.85 \\
& Temporal adjunct & 7.02 & 3.16 & 89.82 & 56.25 & 26.74 & 17.01 \\
& Temporal clause & 9.79 & 7.23 & 82.98 & 43.52 & \textbf{47.22} & 9.26 \\
\hline
\multirow{3}{*}{Embedding Context} 
& Modal & \textbf{10.84} & \textbf{6.38} & 82.79 & 46.14 & \textbf{44.50} & 9.36  \\
& Negation & 9.00 & 6.09 & \textbf{84.91} & 50.72 & 32.97 & 16.30  \\
& Question & 10.03 & 5.47 & 84.50 & \textbf{52.94} & 26.55 & \textbf{20.51}  \\
\hline
\multirow{2}{*}{Scenario Probability} 
& High & \textbf{18.13} & \textbf{8.15} & 73.72 & \textbf{57.50} & 31.16 & 11.35 \\
& Low & 1.87 & 3.80 & \textbf{94.33} & 42.62 & \textbf{37.65} & \textbf{19.73} \\
\hline
\multirow{4}{*}{Political Distance} 
& Left/Right (Linke/AfD) & 4.58 & 4.00 & 91.42 & 45.47 & \textbf{37.22} & 17.31 \\
& Centre-Left/Left (Linke/SPD) & \textbf{19.60} & \textbf{10.74} & 69.66 & 51.76 & 36.38 & 11.85 \\
& Centre-Left/Centre-Right (SPD/CDU) & 11.63 & 4.94 & 83.43 & 50.23 & 30.52 & \textbf{19.25} \\
& Centre-Right/Right (CDU/AfD) & 4.09 & 4.21 & \textbf{91.71} & \textbf{52.61} & 33.57 & 13.82 \\
\hline
\end{tabular}
}
\label{tab:contingency_results}
\end{table*}

We examine the factors that may influence the models' sensitivity to false presuppositions, analyzing our systematically collected experimental conditions.
Since Mistral mostly accommodated misinformation and showed little sensitivity to these factors, we exclude its results from further discussion.
Table \ref{tab:contingency_results} presents the distribution of annotation categories for GPT and LLaMa across each experimental condition. 
An initial inspection reveals distinct patterns in how the two models handle different conditions.
To quantify these differences, we first conducted $\chi^2$-tests to assess the relationship between each annotation label and the four independent conditions: Trigger Type, Embedding Context, Scenario Probability and Political Orientation. 
To control for the family-wise error rate, we applied a Bonferroni correction to all $\chi^2$-tests (adjusted $\alpha = 0.00625$). The reported p-values are corrected.
We then fitted three separate binomial Generalized Linear Models (GLMs) for GPT and LLaMa, each predicting one annotation label against the other two (Accommodation vs. non-Accommodation, Imprecise vs. non-Imprecise, Rejection vs. non-Rejection). The main effects for each model, identified via ANOVA (Type III Wald $\chi^2$ test), are listed in Table \ref{tab:multimodels}.

\paragraph{Presupposition Trigger Type.}

We tested seven different presupposition trigger types to investigate whether these affect the models' ability to detect false presuppositions. 
The Chi-square tests for both LLaMa and GPT showed significant associations between trigger types and overall annotation ($\chi^2_{LLama}(12) = 120.27, p < .001$; $\chi^2_{GPT}(12)= 142.06, p < .001$), with moderate effect sizes ($V_{LLama} = 0.13$; $V_{GPT} = 0.14$).
The Type-III-ANOVA allows a closer look at the prediction criteria for the individual annotation labels (Table \ref{tab:multimodels}). 
The results show that trigger type had a moderate effect on the Imprecise label for GPT ($\chi^2$(3) = 15.24, $p < 0.01$) but no other significant main effects. 
However, interaction effects emerged: In the GLMs predicting the Imprecise label, a strong significant interaction between trigger type and context was observed for LLaMa ($\chi^2$(6) = 22.42, $p < 0.01$), while GPT showed a weaker but still significant effect ($\chi^2$(6) = 13.31, $p < 0.05$).
These results suggest that the models have particular biases toward certain responses depending on the type of trigger. For example: LLama accommodates misinformation more often with interaction particles, while GPT does so with factive verbs (Table \ref{tab:contingency_results}).

\paragraph{Embedding Context.}
We investigated whether the embedding context (question, negation, modal) has an influence on the model's ability to detect false presuppositions. 
For LLaMa, a Chi-Square test of independence showed a significant association with the embedding environment (\(\chi^2(4) = 107.15, p < .001\)), though with a small effect size (\(V = 0.13\)). 
The ANOVA on the binomial GLMs for the individual annotation labels revealed that the embedding context was the strongest main effect of all features for predicting Rejection ($\chi^2$(2)=49.78, $p < 0.001$) 
and Accommodation ($\chi^2$(2)=80.00, $p < 0.001$), 
along with a weak interaction effect with scenario probability on Rejection ($\chi^2$(2)=7.35 $p < 0.05$) (Table \ref{tab:multimodels}). 
In contrast, for GPT, a Chi-Square test of independence showed no significant association ($\chi^2(4) = 3.05,\ p = 1.0$) between embedding context and annotation, suggesting greater robustness to context.
The ANOVA on the binomial GLMs for the single annotation labels confirmed no significant main effect on any of the annotation labels (Table \ref{tab:multimodels}).
Thus, embedding context significantly impacts annotation outcomes for LLaMa, but has no effect on GPT (cf. also Table \ref{tab:contingency_results}). 

\begin{table*}[ht!]
    \caption{$\chi^2$ and p-values of the Anova for 3 Generalized Linear Models (GLMs) per LLM, each predicting one annotation category against the other two (Accommodation vs. non-Accommodation, Imprecise vs. non-Imprecise, Rejection vs. non-Rejection) with the most influential feature highlighted in bold.}
    \centering
    \small
    \resizebox{\textwidth}{!}
    {\begin{tabular}{p{1.7cm} r  | r  l| r  l| r  l | r  l | r  l| r  l}
        \toprule
        \multicolumn{2}{c}{} & \multicolumn{6}{c}{\textbf{GPT}} &  \multicolumn{6}{c}{\textbf{LlaMa}} \\
        \toprule
        \multicolumn{2}{c}{} & \multicolumn{2}{c}{\textbf{Accommodation}} & \multicolumn{2}{c}{\textbf{Imprecise}} & \multicolumn{2}{c}{\textbf{Rejection}}  & \multicolumn{2}{c}{\textbf{Accommodation}} & \multicolumn{2}{c}{\textbf{Imprecise}} & \multicolumn{2}{c}{\textbf{Rejection}} \\
        \toprule
         & Df & $\chi^2$ &  p & $\chi^2$ & p & $\chi^2$ &  p & $\chi^2$  & p& $\chi^2$  & p & $\chi^2$ &  p \\
        \midrule
        
        Trigger Type & 3 & 0.17 & - & 15.24 & \textbf{**} & 5.39 & - &  3.74 & - & 3.02 & - & 4.81 & -   \\
        Context  & 2   & 3.99  & -  & 0.00  & -   & 0.23  & - &   \textbf{80.00} & \textbf{***}  & 2.65  & - & \textbf{49.78}  & \textbf{***} \\  
        Probability & 1   & \textbf{206.28} & \textbf{***} & 29.15  &  \textbf{***} & 25.86  & \textbf{***} & 16.29  &  \textbf{***}  & \textbf{32.93}  & \textbf{***} & 20.25  &  \textbf{***} \\
        Political Distance   & 3 & 164.66 & \textbf{***} & \textbf{46.58}  &  \textbf{***}  & \textbf{30.73}  &  \textbf{***}    & 2.11  & - & 9.87  & \textbf{*}  & 20.92  &  \textbf{***}     \\ 
        \midrule
         
         \multicolumn{14}{c}{Significance codes: \textbf{***} : 0 - 0.001; \textbf{**} : 0.001 - 0.01; \textbf{*} : 0.01 - 0.05;  - : $>$0.05 } \\
         
        \bottomrule
    \end{tabular}
    }
    \label{tab:multimodels}
\end{table*}

\paragraph{Scenario Probability.} 
We tested whether the likelihood of the scenario impacts the models' ability to detect false presuppositions by creating high and low probability scenarios.
As expected, both GPT and LlaMa performed better with low-probability scenarios, showing higher Rejection rates when presuppositions were less realistic (Table \ref{tab:contingency_results}).
However, unexpectedly, LaMa still rejected relatively few low-probability presuppositions, instead accommodating a larger proportion.
Chi-Square tests showed significant associations between scenario probability and annotation categories for both models: \(\chi^2_{LLama}(2) = 86.82, p < .001\) and \(\chi^2_{GPT}(2) = 296.28, p < .001\). Cramér's V indicated moderate effect sizes: \(V_{LLama} = 0.16\) and \(V_{GPT} = 0.29\), with a stronger association in GPT. 
In the Type III ANOVA, scenario probability emerged as the only condition with a strong main effect across all annotation labels for both models (Table \ref{tab:multimodels}).
It significantly influenced Rejection, Imprecise, and Accommodation for both LLaMa (\(\chi^2_{R}(1) = 20.25, p < 0.001\); \(\chi^2_{I}(1) = 32.93, p < 0.001\); \(\chi^2_{A}(1) = 16.29, p < 0.001\)) and GPT (\(\chi^2_{R}(1) = 25.86, p < 0.001\); \(\chi^2_{I}(1) = 29.15, p < 0.001\); \(\chi^2_{A}(1) = 206.28, p < 0.001\)). 
Among these, Accommodation in GPT and Imprecise in LLaMa were the strongest predictors within their respective models. Additionally, scenario probability showed interaction effects with trigger type, as pointed out above.
Thus, scenario probability emerges as a key driver of how each model responds to the false presuppositions.

\paragraph{Political Orientation.}

We examined how the political distance between four German parties (Die LINKE (left), AfD (right), SPD (center-left), and CDU/CSU (center-right)) affects the models' performance in rejecting misinformation. 
Table \ref{tab:contingency_results} shows that the political distance has a clear effect on GPT: it performs best when the right-wing party is included (e.g., AfD - Linke) and worse when parties are politically more similar (e.g., SPD - Linke). LLaMa also is sensitive to political distance, though to a lesser degree.
Chi-square tests indicate a significant relation between political distance and responses for both models: \(\chi^2_{\text{LLaMa}}(6) = 30.38, p < .001\) and \(\chi^2_{\text{GPT}}(6) = 212.09, p < .001\). Cramér's V indicates a moderate effect size for LLaMa (\(V_{\text{LLaMa}} = 0.07\)) and a stronger effect size for GPT (\(V_{\text{GPT}} = 0.18\)). 
The Type III ANOVA further highlights GPT’s sensitivity to political orientation, showing a significant effect across all annotation labels (\(\chi^2_{R}(1) = 30.73, p < .001; \chi^2_{I}(1) = 46.58 , p < .001; \chi^2_{A}(1) = 164.66 , p < .001\)) (Table \ref{tab:multimodels}).
For LLama, political orientation had a moderate but highly significant effect on Rejection (\(\chi^2(3) = 20.92 , p < .001;\) ) and a weaker significant effect on Imprecise (\(\chi^2(3) = 9.87 , p < .005;\)). No significant effect was found for Accommodation. However, a weak interaction effect with Scenario Probability was observed for the Imprecise label in LLaMa (\(\chi^2(3) = 8.3 , p < .005;\)). 
Overall, political distance strongly influences model responses, especially in GPT.

\paragraph{Summary of Results.}
Our findings reveal that Scenario probability is a key factor for both models, with low-probability scenarios increasing rejection rates. Political distance also affects responses, especially in GPT. Embedding context plays a crucial role in LLaMa’s decisions but does not impact GPT.  Trigger type emerges as the fuzziest of all factors: although it hardly appears as a main effect, it does play a subordinate role in a large, unsystematic number of interaction effects.

\section{Discussion}

This paper examined how LLMs handle false presuppositions and, in particular, whether certain linguistic factors contribute systematically to LLMs adopting falsely presupposed information, focusing on political contexts. 
We found that LLMs generally struggle to reject false presuppositions, even when the information is blatantly false. 
Although GPT outperformed the other models, all fell far short of an ideal system that would reject 100\% of false presuppositions. 
This failure is particularly striking given that our prompts were explicitly designed to make rejection an obvious task for a system with basic pragmatic competence. 

Our findings further demonstrate that factors known to shape human presupposition processing – presupposition trigger type, embedding context, scenario probability, and political orientation – also influence LLMs’ sensitivity to false presuppositions, albeit with varying impact across models. While this aligns with psycholinguistic insights, we observe considerable variability in the models’ responses, indicating that they do not exhibit equally systematic patterns. 
These results carry important implications for linguistic analysis and presupposition theory.
On one hand, linguistic presupposition analysis proves valuable for assessing LLMs’ susceptibility to adopt false information. On the other, LLMs may fall short as a tool in presupposition theories unless they exhibit more consistent, human-like patterns.
However, it remains unclear whether LLMs should be expected to mirror human presuppositional behavior at all, given their reliance on statistical text patterns rather than cognitive grounding. Yet this does not preclude further research from leveraging LLMs to shed light on where and why these divergences occur.

Additionally, our findings have important implications for both LLM development and real-world use. 
A key concern is the risk of LLMs spreading misinformation. With 35.6\% of Generation Z using AI tools like ChatGPT for political information \citep{DemokratieStudie2024}, undetected false presuppositions could mislead users and distort public opinion.
Our results suggest that instruction fine-tuning should focus on more than just enriching datasets – models need explicit components to detect the subtle ways misinformation is introduced, especially in political contexts. Improving LLMs' sensitivity to linguistic structures like presupposition triggers and political affiliations could enhance their ability to handle misinformation more effectively.
Another intriguing avenue for future research arises from the strong influence of scenario probabilities and - a little less prominently - party orientation, indicating a reliance on broader context or world knowledge.

\paragraph{Directions for Future Work.}
While this study provides valuable insights into how language models handle false presuppositions, several questions remain. 
Future work could refine the annotation process by analyzing how models express certainty.
For instance, in some cases, the incorrect accommodation of a false presupposition is reinforced with terms like ``indeed" or even supported by hallucinated sources. 
Such cases could provide deeper insights into how models handle misinformation and express certainty.
Additionally, GPT's answers often appeared highly convincing at first glance – its ability to generate lengthy, seemingly coherent responses can mask underlying issues, making errors only apparent upon closer inspection. Its tendency to provide evasive or vague answers further complicates determining whether it has truly rejected a false presupposition. 
Investigating these patterns of vagueness and their impact on user perception could help assess whether models truly reject false presuppositions or simply avoid engaging with them and what it implies for the user's ability to detect their previously uttered misinformation.
Furthermore, so far, no comprehensive study has examined human replies to false presuppositions. Constructing such a dataset could enhance our understanding of the fundamental differences between human and LLM communication.
Lastly, extending the study beyond German (language and political context) could reveal cross-linguistic and cross-cultural differences in handling presuppositions.




\section*{Acknowledgements}
The authors acknowledge financial support by the
project “SAIL: SustAInable Life-cycle of Intelligent Socio-Technical Systems" (Grant ID NW21-059A), an initiative of the Ministry of Culture and Science of the State of Northrhine Westphalia. 

\bibliographystyle{apacite}

\setlength{\bibleftmargin}{.125in}
\setlength{\bibindent}{-\bibleftmargin}

\bibliography{custom}


\end{document}